\documentclass[letterpaper, 10 pt, conference]{ieeeconf}  

\IEEEoverridecommandlockouts                              

\title{\LARGE \bf
Contact-Implicit Trajectory Optimization for Dynamic Object Manipulation
}

\author{Jean-Pierre Sleiman, Jan Carius, Ruben Grandia, Martin Wermelinger, Marco Hutter 
\thanks{This research was supported by the Swiss National Science Foundation through the National Centre of Competence in Digital Fabrication (NCCR dfab), the National Centre of Competence in Research Robotics (NCCR Robotics), and the European Union’s Horizon 2020 research and innovation programme under grant agreement No 780883.}
\thanks{All authors are with the Robotic Systems Lab, ETH Zurich, Zurich, Switzerland.
        {\tt\small jsleiman@ethz.ch}}%
}

\usepackage{xcolor}
\usepackage{graphicx}
\usepackage{hyperref}
\usepackage{amsmath}
\usepackage{cleveref}
\usepackage{amssymb}
\usepackage{bm}
\usepackage{mathtools}
\usepackage{tikz}
\usepackage{epsfig}
\usepackage{epstopdf}
\usepackage{todonotes}
\usepackage{url}
\usepackage{cite}
\usepackage{subfigure}
\crefformat{figure}{#2Fig.~#1#3}
\crefformat{section}{#2Section~#1#3}

\begin{document}

\maketitle
\thispagestyle{empty}
\pagestyle{empty}

\begin{abstract}
We present a reformulation of a contact-implicit optimization (CIO) approach that computes optimal trajectories for rigid-body systems in contact-rich settings.
A hard-contact model is assumed, and the unilateral constraints are imposed in the form of complementarity conditions. 
Newton's impact law is adopted for enhanced physical correctness.
The optimal control problem is formulated as a multi-staged program through a multiple-shooting scheme. This problem structure is exploited within the \textit{FORCES Pro} framework to retrieve optimal motion plans, contact sequences and control inputs with increased computational efficiency.
We investigate our method on a variety of dynamic object manipulation tasks, performed by a six degrees of freedom robot. The dynamic feasibility of the optimal trajectories, as well as the repeatability and accuracy of the task-satisfaction are verified through simulations and real hardware experiments on one of the manipulation problems.      

\end{abstract}

\section{INTRODUCTION}

Over the past years, there has been a significant shift in research attention from industrial robotics towards the development of robots that are capable of a more dexterous and dynamic interaction with their environment. Such an interaction generally involves non-smooth dynamics that can arise from contact constraints, impact events, or frictional forces. Locomotion and dynamic manipulation lie at the forefront of such applications, where analyzing and exploiting the dynamics induced by contacts is fundamental for achieving the desired task at hand. The associated hybrid system dynamics make the design of motion plans and control policies an intricate process. 

 Some of the most prominent control techniques developed to deal with such problems rely on extensions of standard trajectory optimization (TO) methods (direct collocation, single shooting, multiple shooting\dots)~\cite{Betts, Kelly1, Diehl}. These extensions incorporate concepts behind the modelling and simulation of non-smooth contact phenomena and impact dynamics. The primary integration schemes used for the forward simulation of hybrid systems can be identified as the \textit{event-driven} or the \textit{time-stepping (event-capturing) techniques}. The former solves for the different dynamic modes while requiring, at the same time, an event-detection plan that would indicate a transition from the current mode to the next one. On the other hand, time-stepping schemes directly discretize the overall non-smooth dynamics without the need for any detection-step. Adapting these concepts from simulation to control purposes yields two main classes of TO control methods: the \textit{multi-phase approach} and the \textit{contact-implicit approach}, which are analogous to the aforementioned event-driven and time-stepping schemes, respectively.

The multi-phase approach entails an a priori knowledge of the continuous phases or modes, as well as their sequence of occurrence over the full time horizon. The switching times for the discrete transitions can be a result of the optimization program. 
Researchers such as Patterson and Rao~\cite{Rao} have developed a MATLAB program (GPOPS-II) that utilizes the multi-phase strategy as a framework for solving generic hybrid optimal control problems. Other successful applications are repeatedly encountered in the legged-robotics community, such as for bipedal robot locomotion~\cite{Westervelt}, and for quadrupedal locomotion~\cite{Pardo}. However, complications with the use of a predefined event-schedule begin to arise as the number of discrete transitions increases; and that is simply due to the exponential growth of the possible multi-phase sequences with respect to these transitions. 

A remedy to this issue can be found in the mode-invariant approach, also referred to as contact-invariant, contact-implicit, or through-contact approach. The fundamental idea behind it is that, instead of having a predefined mode sequence along with inter-phase constraints, discrete transitions are not given any special treatment; hence, the contact constraints are enforced at every single grid-point. The method is also capable of incorporating discontinuous jumps in the states -- such as in velocities during impact events -- by treating the whole problem on an impulse level, thereby not differentiating between finite forces and impulsive forces. 

Both multi-phase and contact-implicit schemes can be implemented along with a variety of underlying design choices. To elaborate, one could flexibly choose any of the direct TO methods, with either a soft-contact model or a hard-contact model assumption; and in case the latter was chosen, one would have to deal with the resulting set-valued force laws by either resorting to an augmented-Lagrangian approach~\cite{Glocker} or by solving a linear complementarity problem (LCP)~\cite{Trinkle}. Furthermore, one could decide to add the contact forces into the vector of decision variables, or to resolve them separately from the optimization iterations.   

Some remarkable work has been done in the computer graphics and the robotics communities on optimal motion planning and control, under the framework of contact-implicit optimization (CIO):
For instance, Mordatch et al.~\cite{Mordatch,Mordatch2,Mordatch3} use a contact-invariant approach for producing physics-based animations of anthropomorphic, contact-rich tasks like walking, climbing, crawling, moving objects, and multi-fingered manipulation. Although some variations of their original CIO approach are implemented, there is a common fundamental treatment of the contact
constraints, which are imposed through penalty terms added to the objective function (soft-constraints); and this relaxation, while it certainly aids the gradient-based solver in efficiently converging to a local minimum, it also tolerates non-physical behavior caused
by allowing contact forces to act at a distance. Another soft-constrained
CIO formulation is achieved in~\cite{Tassa} by a differential-dynamic-programming (DDP) optimal control method, where the
hard unilateral contact constraints and frictional forces are dealt with by solving
a stochastic linear complementarity problem rather than an LCP. With
this, they get rid of the discontinuous dynamics and are left with a continuously differentiable system, while retrieving an optimal controller that is also robust against
uncertainties. Neunert et al.~\cite{Neunert} have developed a nonlinear model predictive control (NMPC) algorithm that solves, online and at high rates, the nonlinear optimal control problem using a generalization of the iLQR method. The NMPC scheme is tested on a quadruped, while adopting a contact-implicit formulation which assumes a soft-contact model with smoothing elements. On the other hand, Carius et al.~\cite{Carius} rely on the augmented-Lagrangian approach, specifically the Gauss-Seidel method combined with
a proximal point projection (SORprox), in order to solve the set-valued force laws arising from the hard frictional contacts. This is performed as part of a Moreau time-stepping scheme which is used to simulate the non-smooth dynamics within an iterative-LQR (iLQR) optimal control plan.

In this paper, we propose a variation on the CIO formulation introduced by Posa et al.~\cite{Posa}. Their method is originally motivated by the Stewart and Trinkle time-stepping scheme \cite{Trinkle}, where multi-contact dynamics are formulated as a linear complementarity problem (LCP). They make use of the simultaneous nature of direct collocation by incorporating the contact forces as optimization variables, while additionally imposing the unilateral hard-contact constraints in the form of complementarity conditions. In fact, the transcription of hybrid optimal control problems into finite-dimensional optimization programs including complementarity conditions has been initially proposed by Yunt $\&$ Glocker~\cite{Yunt}; but unlike~\cite{Posa}, the optimal control inputs are retrieved separately from the contact forces and states through a bilevel optimization. We follow the same fundamental steps, while reformulating the nonlinear program (NLP) within the multiple-shooting framework instead. This modification is not immediately straightforward as it sacrifices the method's physical correctness, unless some extra steps associated with the introduction of an impact-law are considered. The new NLP-structure enables us to exploit the computational power of the \textit{FORCES Pro}~\cite{Forces1} commercial tool, which generates highly customized and efficient optimization solvers based on the primal-dual interior point algorithm. Its effectiveness is mainly a result of an efficient linear-system solver that exploits the multi-staged problem-structure, in order to solve the resulting Karush-Kuhn-Tucker (KKT) system \cite{Forces2}. we demonstrate the use of this method for solving dynamic object manipulation problems similar to that tackled in~\cite{CIOblock}. Unlike~\cite{CIOblock}, wherein Posa's CIO scheme is also tested, we achieve optimal state-trajectories -- at a significantly lower computational time -- that are verified to be dynamically feasible through simulation as well as hardware experimentation. In fact, relatively simple examples in \cite{Posa} are solved within a few minutes, compared to average convergence times ranging between 50 to 800 milliseconds for our setups. The resulting motion plans are shown to adhere to the imposed contact conditions, and result in realistic contact forces. These indeed form the crucial ingredients for successfully attaining the desired manipulation-task goal, reliably and accurately, without the need for any sort of post-optimization modifications.

\begin{figure}[t!]
\centering
\vspace*{0.25cm}
\begin{minipage}{0.45\textwidth}
\centering
\begin{subfigure}[]{
   \includegraphics[width = 6.25 cm, keepaspectratio]{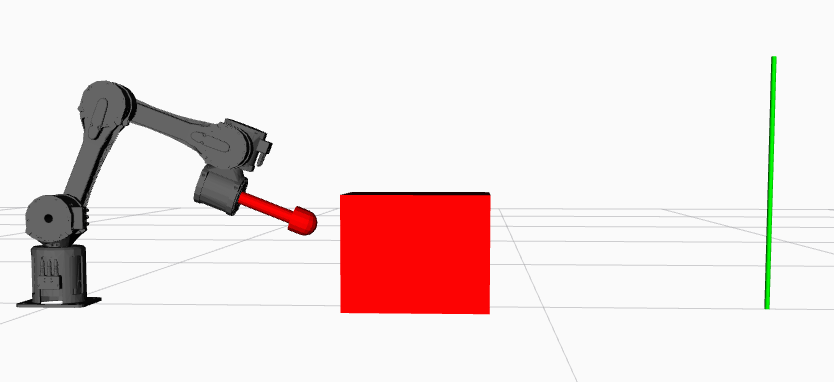}
 }
\end{subfigure}
\end{minipage} 
\begin{minipage}{0.45\textwidth}
\centering
\begin{subfigure}[]{
   \includegraphics[width = 6.25 cm, keepaspectratio]{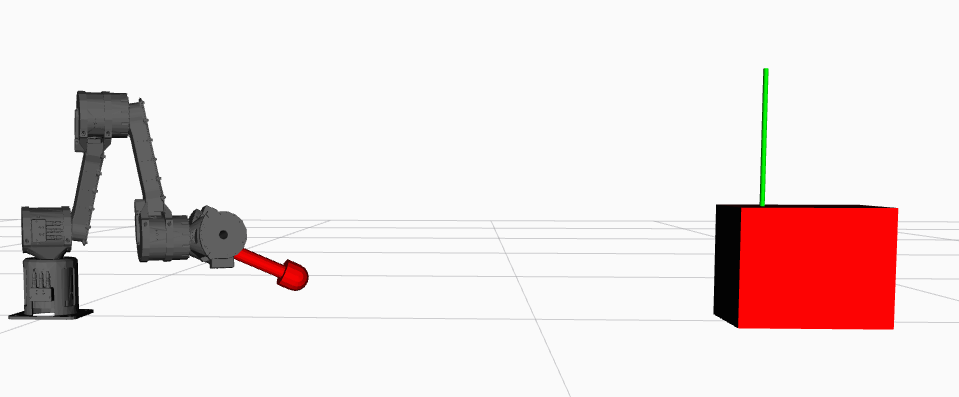}
 }
\end{subfigure}
\end{minipage} 
\caption{\textbf{(a)} The initial states of the robot and the block are defined with zero velocities and pre-defined configurations. The manipulator is required to dynamically push the block such that \textbf{(b)} the final desired configuration of the block is attained and the robot comes at rest with an arbitrary configuration by the end of the manipulation period}
\label{fig:CentralGoal}
\end{figure}
\begin{figure}[b!]
\centering
\begin{subfigure}[]{
   \includegraphics[height = 2.5cm, keepaspectratio]{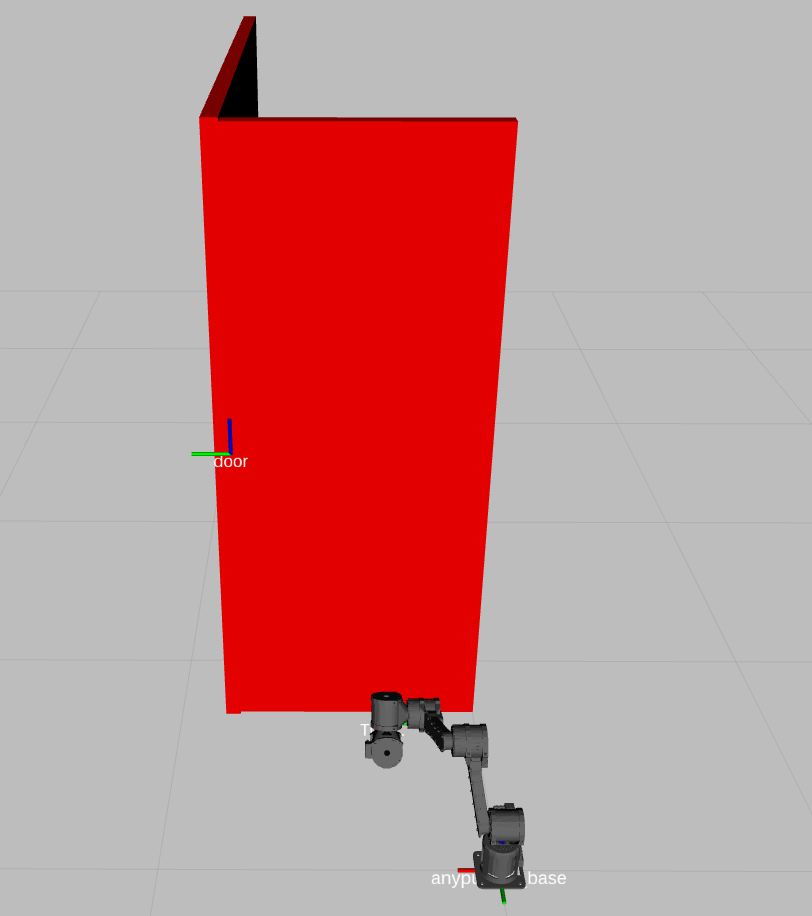}
    \label{ANYpulatorDoor}
}
\end{subfigure}
\begin{subfigure}[]{
\includegraphics[height = 2.5cm, keepaspectratio]{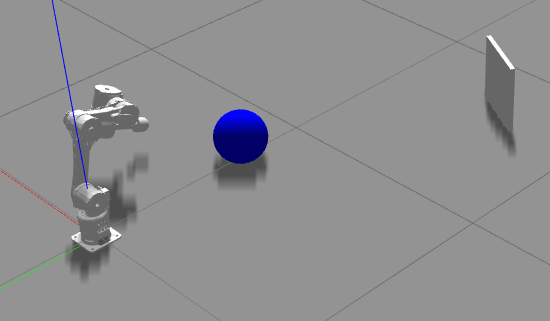}
\label{ANYpulatorBall}
}
\end{subfigure}
\caption{\textbf{(a)} Robot Manipulator dynamically pushing a door open and \textbf{(b)} pushing a ball to a desired goal }
\label{fig:DoorBall}
\end{figure}
\section{PROBLEM DESCRIPTION}

\subsection{Dynamic Object Manipulation}
Dynamic manipulation belongs to the family of non-prehensile (graspless) manipulation, which generally includes tasks such as pushing, tumbling, throwing, and catching~\cite{Lynch, DynamicManipulation}. Essentially, what makes the task ``dynamic", is that, unlike the quasi-static case, inertial effects have to be taken into account due to the relatively fast motions; and secondly, it is possible for the manipulator to lose contact with the object during the assigned manipulation period. This, in turn, makes it mandatory for one to reason about the dynamics induced by the environment on the uncontrollable object, while planning and controlling the robot motion that would satisfy the objective. 

We consider a variety of dynamic pushing problems in this work. Ultimately, our goal is to formulate an NLP that would allow a six degrees of freedom (DOF) robot manipulator (\cref{sec:ANYpulator}) to dynamically push a block from a given initial state, to a final desired position that could lie outside the robot's workspace (see~\cref{fig:CentralGoal}). The problem formulation is general enough to work on diverse manipulation problems as well, such as pushing a door open, or pushing a ball into a desired direction and up to an assigned goal (see~\cref{fig:DoorBall}).
We elaborate on the problem specifics and results of the block-task, as it comprises all relevant effects from a modelling and computational perspective, due to the action of the environment on the block in the form of dry-friction.
\begin{figure}[t!]
\centering
\vspace*{0.5cm}
\begin{subfigure}[]{
   \includegraphics[height = 4 cm, keepaspectratio]{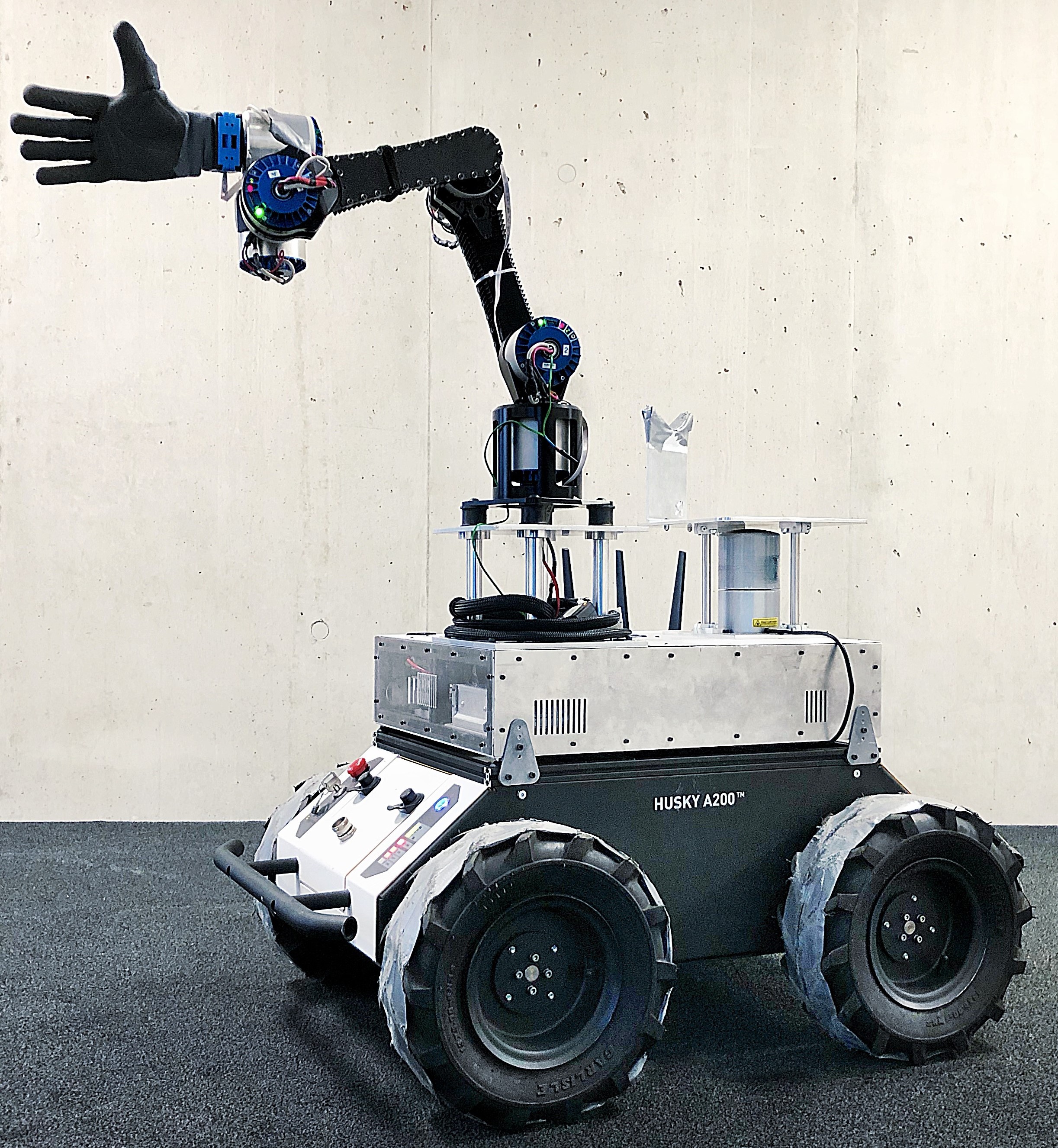}
    \label{fig:mANYpulator}
 }
 \vspace*{0.5cm}
\end{subfigure}
\begin{subfigure}[]{
\includegraphics[height = 4 cm, keepaspectratio]{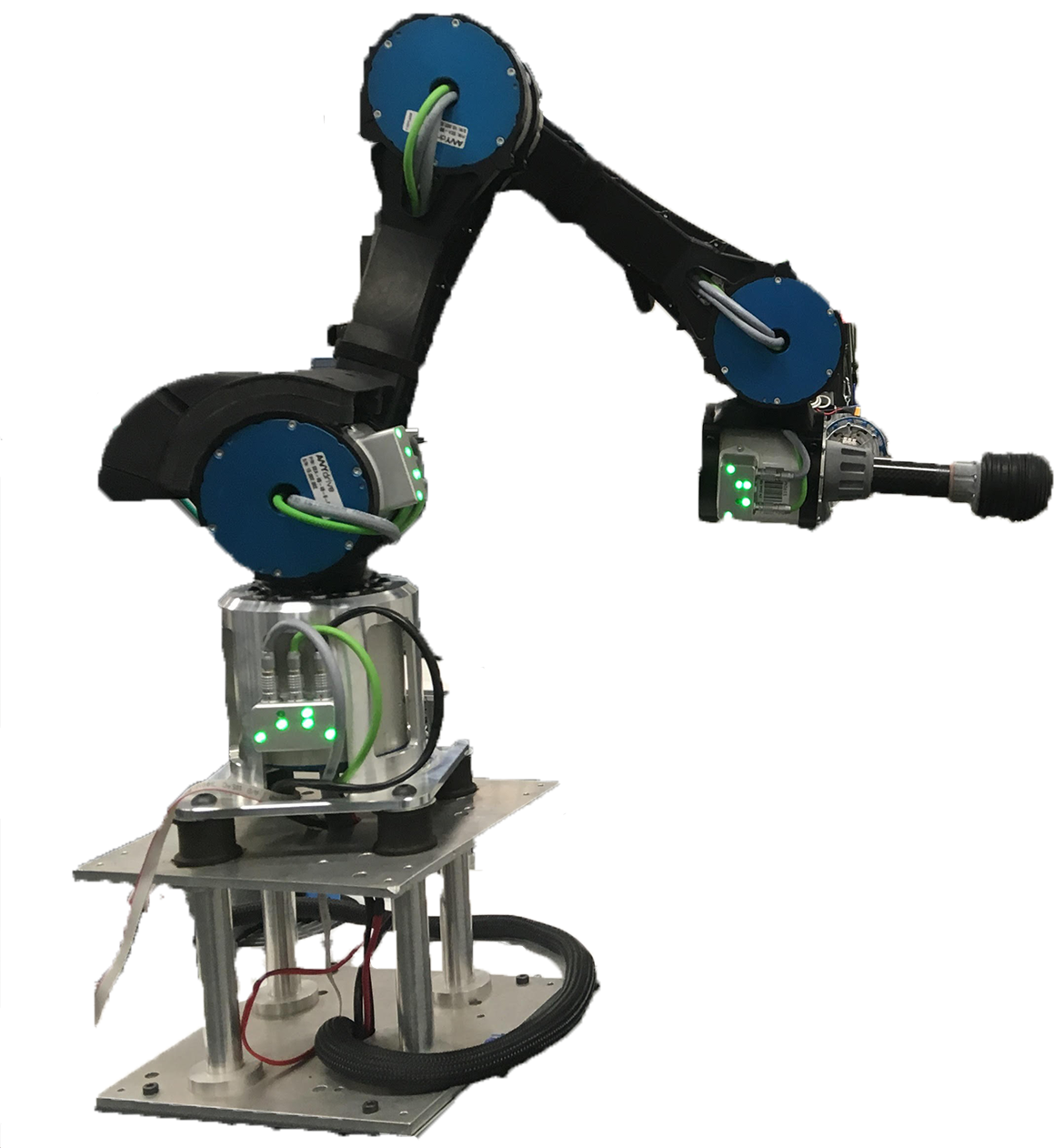}
\label{fig:MyANYpulator}
 }
\end{subfigure}
 \vspace*{-2mm}
\caption{\textbf{(a)} ANYpulator robotic arm mounted on top of a Husky mobile platform \cite{anypulator2}. The series elastic actuated joints allow for highly dynamic motions, and high impact robustness. \textbf{(b)} ANYpulator with a custom-built end-effector that is used in all ANYpulator-Object manipulation problems. The geometrical structure of the end-effector allows us to neglect the last joint's states and input-torque}
\label{fig:GeneralAnypulator}
\end{figure}

\subsection{ANYpulator} \label{sec:ANYpulator}

We test the proposed framework with ANYpulator~\cite{anypulator1,anypulator2}, a 6-DOF robot arm depicted in~\cref{fig:GeneralAnypulator}. It comprises a set of robust and mechanically compliant series-elastic actuators (ANYdrive \cite{ANYdrive}) that feature precise position and torque control as well as high impact robustness. The six joints are responsible for: Shoulder Rotation (${SH_{ROT}}$), Shoulder Flexion/Extension (${SH_{FLE}}$), Elbow Flexion/Extension (${EL_{FLE}}$), Wrist Flexion/Extension (${WR_{FLE}}$), Wrist Deviation (${WR_{DEV}}$), and Wrist Pronation (${WR_{PRO}}$).  
For the sake of our dynamic pushing applications, a custom-built end-effector is used for all the object-manipulation problems. The specialized tool basically constitutes of a cylindrical rod with a spherical head, as shown in~\cref{fig:MyANYpulator}. Consequently, the rotation of the last joint would have no contribution to the resulting solutions of the applications at hand, thus allowing us to extract a 5-DOF dynamic model from the manipulator equations of motion. Such a consideration helps decrease the optimization program-size significantly, which in turn reduces the associated computational effort.

\section{Methodology}

Tackling the aforementioned manipulation problems with the contact-implicit approach requires an accurate dynamic model of the robot and object.
We consider a frictionless end-effector to object contact in the following derivations, since the tangential components of the contact forces/impulses only play a minor role in our dynamic pushing tasks. This assumption helps reduce computational time at the expense of a slight loss in physical accuracy.

\subsection{Model Formulation} 

Noting that in our applications the generalized velocities ${\textbf{u} \in \mathbb{R}^{n_u}}$ coincide with the time-derivatives of the generalized positions ${\textbf{q} \in \mathbb{R}^{n_q}}$, the fully-actuated dynamics of the robot are concatenated with the non-actuated dynamics of the object to form the following underactuated rigid body dynamic equations: \\
\begin{equation} \label{EOM}
    \resizebox{0.91\hsize}{!}{
        $\underbrace{\begin{bmatrix}
        \mathbf{M}_r(\mathbf{q}_r) & \mathbf{0}_{5\times n_{q_o}} \\
        \mathbf{0}_{n_{q_o}\times 5} & \mathbf{M}_o(\mathbf{q}_o)
        \end{bmatrix}}_{\mathbf{M(\mathbf{q})}} 
        \underbrace{\begin{bmatrix}
        \Ddot{\mathbf{q}}_r \\
        \Ddot{\mathbf{q}}_o
        \end{bmatrix}}_{\Ddot{\mathbf{q}}}
        +
        \underbrace{\begin{bmatrix}
        \mathbf{h}_r(\mathbf{q}_r,\dot{\mathbf{q}}_r) \\
        \mathbf{h}_o(\mathbf{q}_o,\dot{\mathbf{q}}_o)
        \end{bmatrix}}_{\mathbf{h}(\mathbf{q},\dot{\mathbf{q}})}
        = 
        \mathbf{S}^T\pmb{\tau}  + 
        \mathbf{J}^T_N(\mathbf{q}) \pmb{\lambda}_N
        +
        \begin{bmatrix}
        \mathbf{0}_{5\times1} \\
        \mathbf{F}_{ext}
        \end{bmatrix}$
        }
\end{equation} 
as well as the following impulse-momentum equations:
\begin{equation} \label{IME}
    \mathbf{M}(\mathbf{q}) (\dot{\mathbf{q}}^+ - \dot{\mathbf{q}}^-) = \mathbf{J}^T_N(\mathbf{q})\pmb{\Lambda}_N + \begin{bmatrix}
        \mathbf{0}_{5\times1} \\
        \pmb{{\mathcal{F}}}_{ext}
        \end{bmatrix} 
\end{equation}
where the subscripts \textit{r} and \textit{o} refer to quantities corresponding to the robot and the object, respectively. Letting ${n_q = 5+n_{q_o}}$, ${\mathbf{M} \in \mathbb{R}^{n_q\times n_q}}$ is the generalized mass matrix, while ${\mathbf{h} \in \mathbb{R}^{n_q}}$ contains Coriolis and centrifugal terms, as well as gravitational and viscous damping terms. The selection matrix ${\mathbf{S}^T = \begin{bmatrix}\mathbb{I}_{5\times 5} \ \ \ \mathbf{0}_{5\times n_{q_o}}\end{bmatrix}^T \in \mathbb{R}^{n_q\times 5}}$ maps joint input torques to generalized forces, while ${\mathbf{F}_{ext} \in \mathbb{R}^{n_{q_o}}}$ represents the generalized forces applied by the environment on the manipulated object, such as dry friction.
We define the gap function ${\pmb{\phi}(\mathbf{q})}$, which is the signed distance between potentially colliding bodies. In our case, ${\phi(\mathbf{q})}$ is a scalar-valued function indicating the minimum distance between the robot's end-effector and the object. Consequently, the normal Jacobian ${\mathbf{J}_N = {\partial \phi(\mathbf{q})}/{\partial \mathbf{q}} \in \mathbb{R}^{1\times n_q}}$ and the magnitude of the normal contact force ${\lambda_N \in \mathbb{R}_{\geq 0}}$ are defined accordingly. Moreover, the post- and pre-impact generalized velocities are given by ${\dot{\mathbf{q}}^+}$ and ${\dot{\mathbf{q}}^-}$ in (\ref{IME}), while ${\Lambda_N \in \mathbb{R}_{\geq0}}$ and ${\pmb{{\mathcal{F}}}_{ext}}$ are the normal contact impulse-magnitude and external impulse, respectively.  

\subsection{Nonlinear Program Setup} \label{subsec:NLP}

The generic finite-dimensional optimization program is defined using the following multi-staged framework (compatible with that of \textit{FORCES Pro's} high-level interface): \\
\begin{equation} \label{forces}
\begin{cases}
    \underset{\mathbf{z}_1,...,\mathbf{z}_N}{\min} \hspace{0.5cm} F_N(\mathbf{z}_N) + \sum\limits_{k = 1}^{N-1}F(\mathbf{z}_k) \\\\
    s.t. \hspace{0.5cm} \mathbf{E}_k \ \mathbf{z}_{k+1} = \mathbf{c}(\mathbf{z}_k) \hspace{1.15cm} \forall \ k = 1,...,N-1 \\ \vspace{-0.3cm}\\
    \hspace{1.1cm} \mathbf{S}_1 \ \mathbf{z}_1 = \mathbf{z}_{init} \\
    \hspace{1.1cm} \mathbf{S}_N \ \mathbf{z}_N = \mathbf{z}_{final} \\ \vspace{-0.3cm} \\
    \hspace{1.1cm} \underbar{$\mathbf{z}$}_k \leq \mathbf{z}_k \leq \Bar{\mathbf{z}}_{k} \hspace{1.6cm}  \forall\  k = 1,...,N \\
    \hspace{1.1cm} \underbar{$\mathbf{h}$}_k \leq \mathbf{h}(\mathbf{z}_k) \leq \Bar{\mathbf{h}}_{k} \hspace{1cm}  \forall \ k = 1,...,N \\
\end{cases}
\end{equation} 
where ${\mathbf{z}_k}$ is the vector of optimization variables at stage $k$; $F$ and $F_N$ are the inter-stage cost and the terminal cost functions, respectively. ${\mathbf{E}_k}$, ${\mathbf{S}_1}$, ${\mathbf{S}_N}$ are selection matrices and ${\mathbf{c}(\mathbf{z}_k)}$ is a state-transition function which maps the states from stage ${k}$ to stage ${k+1}$; it could be given as a result of any desired integration scheme, but to be more general, this function could also consist of other nonlinear equality constraints.

With the adopted CIO approach, we formulate and solve a discrete-time optimal control problem using the multiple-shooting method, along with a set of complementarity constraints. These are added to ensure that the unilateral contact conditions hold at every discrete grid-point. Such a setup results in a so-called mathematical program with complementarity constraints (MPCC). 

We can now indicate and discuss the various components required to build our nonlinear program. Most of the upcoming considerations are generally applied to solve all of our manipulation tasks; nonetheless, there are still some problem-specifics associated with the object's geometry and its configuration-space topology. As previously mentioned, the emphasis will be made on the ANYpulator-Block manipulation problem. \\

\textit{Transcription parameters:} When choosing the discretization parameters, there is an inherent trade-off between computational speed and integration accuracy. We attain a fair compromise by using a time-horizon of ${T=1.5\ \text{s}}$ and setting the number of stages ${N = 40}$; hence, leaving us with a time-step size of ${{\Delta t=T}/{(N-1)}=0.03846 \ \text{s}}$. \\
    
\textit{State-transition function:} The forward simulation of the dynamics is carried out on the basis of a semi-implicit Euler scheme. Similar to the classical explicit Euler method, it is a first-order integrator; however, it additionally holds inherent energy-conservation properties that make it more reliable. Therefore, the nonlinear equality constraints responsible for satisfying the dynamics are given as follows:
    \begin{equation} \label{StateTransition}
    \resizebox{0.99\hsize}{!}{$
    \begin{cases}
        \dot{\mathbf{q}}_{k+1} = \dot{\mathbf{q}}_{k} + \Delta t\cdot\mathbf{M}^{-1}_k\left(\mathbf{S}^T\pmb{\tau}_k - \mathbf{h}_k
        +
         \begin{bmatrix}
        \mathbf{0}_{5\times1} \\
        \pmb{\lambda}_{f_k}
        \end{bmatrix} + 
        \mathbf{J}_{N_k}^T \lambda_k\right) \\
        {\mathbf{q}}_{k+1} = \mathbf{q}_k + \Delta t\cdot\dot{\mathbf{q}}_{k+1}
    \end{cases}$}
    \end{equation} 

It is important to note the difference between $\lambda$, $\pmb{\lambda}_f$ appearing in (\ref{StateTransition}) and $\lambda_N$, $\mathbf{F}_{ext}$ in (\ref{EOM}). The latter quantities are infinite during an impact, unlike $\lambda$ and $\pmb{\lambda}_f$ which are defined in relation to the normal percussion $P_N$ and frictional percussion $\mathbf{P}_f$ -- which arise as a result of discretizing the equality of measures \cite{Glocker} -- as follows: 
\begin{equation}
    \begin{split}
        P_N &= \Delta t \cdot\lambda  \\
        \mathbf{P}_f &= \Delta t \cdot \pmb{\lambda}_f
    \end{split}
\end{equation}
where $P_N$ is defined in terms of $\lambda_N$ and $\Lambda_N$, while $\mathbf{P}_f$ in terms of $\mathbf{F}_{ext}$ and $\pmb{\mathcal{F}}_{ext}$:
\begin{equation}
\begin{split}
    d\mathbf{P}_{f} \ \ \ &= \ \ \ \ \ \ \ \mathbf{F}_{ext}\ dt \ \ \ \ \ + \ \ \ \ \pmb{\mathcal{F}}_{ext} d\eta \\
    dP_{N} \ \ \ &= \underbrace{\lambda_{N}\ dt}_{\text{Lebesgue integrable term}}  + \ \ \ \underbrace{\Lambda_{N}d\eta}_{\text{atomic term}} 
\end{split}
\end{equation} \\
$d\eta$ is such that $\displaystyle\int d\eta = 1$ at the time of a discontinuity and is zero elsewhere.  
From here on out, the words \textit{contact force} and \textit{frictional force} refer to the quantities introduced in (\ref{StateTransition}) and not (\ref{EOM}), unless otherwise stated. \\  

\textit{Optimization variables:} The vector of optimization variables is given by the robot's joint input torques, the system's states, the normal contact force $\lambda$, and the dry friction force acting on the block by the environment. We define ${\pmb{\lambda}_f = -_\mathcal{I}\hat{\mathbf{n}} \cdot F_f}$ where ${_\mathcal{I}\hat{\mathbf{n}}}$ is a unit vector indicating the assigned direction of motion for the block, expressed in the robot base frame, which is also used as our inertial frame of reference.
\noindent Therefore, ${\mathbf{z}_k = \begin{bmatrix}\pmb{\tau}^T & \mathbf{x}^T & \lambda & F_f \end{bmatrix}^T_k}$ where ${\pmb{\tau}_k \in \mathbb{R}^5}$, ${\mathbf{x}_k=\begin{bmatrix}\mathbf{q}^T & \dot{\mathbf{q}}^T \end{bmatrix}^T_k=\begin{bmatrix}\mathbf{q}^T_r & \mathbf{q}^T_o & \dot{\mathbf{q}}^T_r & \dot{\mathbf{q}}^T_o \end{bmatrix}^T_k \in \mathbb{R}^{(10+2n_{q_o})}}$, and $\lambda_k$, ${F_{f_k} \in \mathbb{R}_{\geq 0}}$. \\

\textit{Upper/Lower bounds:} The maximum joint velocity is set to ${\dot{q}_{{max}} = 12 \ \text{rad/s}}$, while the maximum allowable torque as ${\tau_{max} = 40 \ \text{N.m}}$. The joint position limits are also imposed in the program. \\

\textit{Boundary conditions:} Initial boundary conditions are used to specify the starting configuration of both the manipulator and the object, while the satisfaction of the assigned manipulation task is ensured through final boundary conditions. It is worth noting that in the case of the block, the configuration variables are ${\mathbf{q}_o = \begin{bmatrix} x_b & y_b & \theta_b \end{bmatrix}^T}$, and the associated manipulation task involves specifying values for $\Delta L_{des}$ and ${\theta_{b_{init}} = \theta_{b_{des}}}$ to indicate the desired displacement and direction of travel, respectively. This is done while also requiring that the block does not rotate at all throughout the full time-horizon. \\

\textit{Objective function:} A common objective function that aims at minimizing energy and also puts a cost on the wrist joints velocities, is used. The latter term is introduced because it could be noticed that an unnecessary movement of the two wrist joints occurs if we only minimize energy, as their motion does not contribute much to the total energy expenditure. Furthermore, the torques and joint velocities are normalized as they do not have the same units. Therefore, the normalized cost function is given as follows:
\begin{equation}
        J(\mathbf{z}_k) = \displaystyle \sum \limits_{k=1}^N \Delta t\cdot\left(\dfrac{\pmb{\tau}_k^T \mathbf{R} \pmb{\tau}_k}{\tau_{max}^2} +  \dfrac{\dot{\mathbf{q}}_{r_k}^T \mathbf{Q} \dot{\mathbf{q}}_{r_k}}{\dot{q}^2_{max}}\right)
    \end{equation}
    where $\mathbf{R}$ and $\mathbf{Q}$ are weighting matrices. \\

    \textit{Complementarity conditions:} The following
    constraints ensure no penetration between colliding bodies, no forces acting at a distance, and only allow forces that can push and not pull: 
    \begin{equation} \label{ComplementarityConstraints}
        0\leq \phi_k \perp \lambda_k \geq 0 \implies 
        \begin{cases}
        \phi_k \geq 0 \\
        \lambda_k\geq 0 \\
        \phi_k \cdot \lambda_k = 0
        \end{cases}
    \end{equation}
    A subtle yet fundamental difference between the formulation in this paper and \cite{Posa} can be seen in the relative indexing of $\phi$ and $\lambda$ in (\ref{StateTransition}) compared to (\ref{ComplementarityConstraints}). To elaborate, in \cite{Posa} the state-transition function was obtained by applying a zero-order-hold (z.o.h) on $\lambda_{k+1}$ instead of $\lambda_k$. Consequently, \textit{``the contact force over the whole time-step can be non-zero if and only if the distance function is zero at the \textbf{end} of the interval"} \cite{Posa}. Whereas
here, the condition holds if and only if the gap is closed at the \textbf{start} of
the interval. This is indeed problematic since there is no more guarantee that the contact event
$\phi_k = 0$ is held over the entirety of the time-step, even though the contact
force keeps acting unjustifiably at a non-zero distance and would only vanish at
the end of it. Moreover, the complementarity conditions, which are purely imposed on a position-level, only ensure that the resulting normal contact force is large enough to prevent penetration in
the next step. Nevertheless, there is certainly no upper limit on how big this force
can be, thus indicating that it is quite arbitrary and merely directed towards satisfying the position-based conditions. To resolve this issue, we resort to the theory underlying the simulation of dynamics involving hard contacts and impact events, using the time-stepping technique. What we extract from that is the need for an impact law, such as Newton's law of restitution: 
\begin{equation} \label{restitution}
    \phi_k = 0 \ : \ \ \ \ \ \ \ 0 \leq P_{N_k} \perp \left(\gamma_{k+1} + \epsilon \gamma_k \right) \geq 0
\end{equation}
where the constant $\epsilon$ is Newton's coefficient of restitution, while $\gamma$ is the relative separation velocity. Noting that (\ref{restitution}) includes the case of a non-participating (superfluous) contact \cite{Glocker2}, which does not occur in any of our applications, we can reduce this expression to:
\begin{equation}
\begin{split}
    P_{N_k}\cdot\left(\gamma_{k+1} + \epsilon\gamma_k\right) = 0 \\ 
   \Rightarrow \ \Delta t\cdot \lambda_k \cdot\left(\dot{\phi}_{k+1} + \epsilon\dot{\phi}_k\right) = 0 \\ 
    \Rightarrow \ \Delta t\cdot \lambda_k \cdot\left(\mathbf{J}_N(\mathbf{q}_{k+1})\dot{\mathbf{q}}_{k+1} + \epsilon\mathbf{J}_N(\mathbf{q}_{k})\dot{\mathbf{q}}_k\right) = 0
\end{split}
\end{equation}
where the states at the end of the interval $\mathbf{q}_{k+1}$ and $\dot{\mathbf{q}}_{k+1}$ can be retrieved from (\ref{StateTransition}). A restitution coefficient of $\epsilon = 0$ is adopted to ensure a purely inelastic collision event, thus effectively imposing a no-rebound condition.

We assume a static Coulomb model to describe the frictional force arising between the plane and the block 
\begin{equation} \label{eq:staticFriction}
    \begin{cases}
    F_f = \lambda \ \ \ \ \text{if} \ v_{dir} = 0 \ \text{and} \ \lambda < F_s \\
    F_f = F_s \ \ \ \text{if} \ v_{dir} > 0
    \end{cases}
\end{equation} 
where $v_{dir}$ is a positive scalar quantity (unidirectional displacement due to pushing and no pulling) that signifies the velocity along the desired direction of motion ${v_{dir}= _\mathcal{I}\hat{\mathbf{n}}^T \dot{\mathbf{q}}_o}$ and $F_s$ is the static friction force or the breakaway force $F_s = \mu_s N_f$.
For the sake of avoiding the conditional if-statements appearing in (\ref{eq:staticFriction}), we introduce the following constraints instead: \\
\begin{equation} \label{MyFriction}
    \begin{cases}
    \begin{rcases}
    v_{dir}\left(F_s - F_f\right) = 0 \\
        F_s-F_f \geq 0\\
            v_{dir} \geq 0\\
     \end{rcases}
    \iff 0\leq v_{dir} \perp \left(F_s - F_f\right) \geq 0 \\
    (F_s-F_f)(\lambda - F_f) = 0
    \end{cases}
\end{equation} 

\textit{Generalized distance function:}
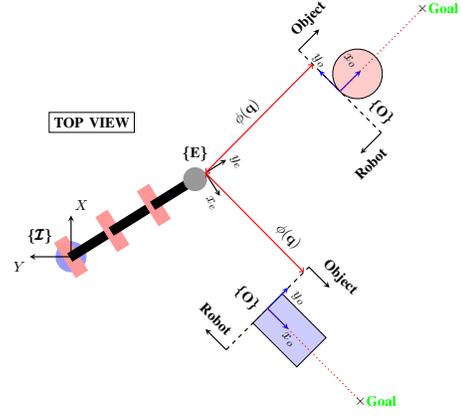
\begin{figure}
    \centering
    \scalebox{0.55}{
\begin{tikzpicture}
\def\iangle{atan(5/5)};
\def\i2angle{atan(8/5)};
\def\icangle{atan(5/8)};
\coordinate (O) at (1.25, 2.25);
\coordinate (A) at (3.25, 4.25);
\draw [dashed,thick](O)-- (A);
 \node[draw, circle, minimum size=1.2cm, rotate=-\iangle, yshift=0.6cm,fill = red!20] at (4,8.5) {};
\node[draw, rectangle, minimum height =1.5cm, minimum width = 1cm, rotate=\iangle, yshift=-0.75cm,fill = blue!20] at (2.25,3.25) {};

\draw [thick,-stealth](A) -- (3.75,3.75);
\draw [thick,-stealth](O) -- (0.75,2.75);

\node[rotate=-\iangle] at (4, 4) {\textbf{Object}};

\node[rotate=-\iangle] at (1,3) {\textbf{Robot}};

\draw [thick,-stealth,blue](4,8.5) -- (3.5,9) node (v3) {};
\draw [thick,-stealth,blue](4,8.5) -- (4.5,9);

\draw [dashed,thick](3,9.5) node (v4) {} -- (5,7.5) node (v5) {} ;

\node[rotate=-\iangle]  at (3.5,9.25) {$y_o$};
\node[rotate=-\iangle]  at (4.25,9.25) {$x_o$};
\node[font=\bf,rotate=-\iangle] at (5,8) {\{O\}};

\draw [thick,-stealth,blue](2.25,3.25) -- (2.75,2.75);
\draw [thick,-stealth,blue](2.25,3.25) -- (2.75,3.75);

\node[rotate=-\iangle]  at (2.75,2.5) {$x_o$};
\node[rotate=-\iangle]  at (3,3.5) {$y_o$};
\node[font=\bf,rotate=-\iangle] at (1.75,3.5) {\{O\}};

\node[fill, circle, minimum size=0.75cm, rotate=\iangle,fill = blue!40] (v1) at (-2.5,4.5) {};
\node[fill, rectangle, minimum height =0.35cm, minimum width = 1cm, rotate=-\i2angle,fill = red!40] (v1) at (-2.5,4.5) {};
\node[fill, rectangle, minimum height =0.05cm, minimum width = 1.25cm, rotate=\icangle,fill = black] at (-2,4.8125) {};

\node[fill, rectangle, minimum height =0.35cm, minimum width = 1cm, rotate=-\i2angle,fill = red!40] at (-1.5,5.125) {};
\node[fill, rectangle, minimum height =0.05cm, minimum width = 1.25cm, rotate=\icangle,fill = black] at (-1,5.4375) {};

\node[fill, rectangle, minimum height =0.35cm, minimum width = 1cm, rotate=-\i2angle,fill = red!40] at (-0.5,5.75) {};
\node[fill, rectangle, minimum height =0.05cm, minimum width = 1.25cm, rotate=\icangle,fill = black] at (0,6.0625) {};

\draw[thick,-stealth] (0.5,6.375) -- (1.25,6.84375);

\node[fill, circle, minimum size=0.58cm, rotate=-\i2angle,fill = black!40] (v2) at (0.5,6.375) {};

\draw [thick,-stealth,black](-2.5,4.5) -- (-2.5,5.5);
\draw [thick,-stealth,black](-2.5,4.5) -- (-3.5,4.5);
\node at (-2.25,5.75) {$X$};
\node at (-3.75,4.25) {$Y$};
\node[font=\bf] at (-3.25,5) {\{$\pmb{\mathcal{I}}$\}};

\draw [thick,-stealth,black](0.75,6.53125) -- (1.1,5.97125) ;
\node[ rotate=-\i2angle] at (1.5,6.8) {$y_e$};
\node[ rotate=-\i2angle]  at (0.9,5.8) {$x_e$};
\node[font=\bf] at (0.5,7) {\{E\}};
\node[draw, rectangle,font=\bf] at (-2,7.75) {TOP VIEW};

\draw [thick,<->,red](0.75,6.53125) -- (3.125,4.125) ;

\draw [thick,<->,red](0.75,6.53125) -- (3.3438,9.1562) ;

\node[rotate=-\iangle] at (2.75,5) {$\phi(\mathbf{q})$};

\draw[thick,-stealth,black] (3,9.5) -- (3.5,10);
\draw[thick,-stealth,black] (5,7.5) -- (4.5,7);

\draw [dotted,thick,red] (4,8.5) -- (6,10.5);
\node[] at (6,10.5) {$\times$};

\draw [dotted,thick,red] (2.25,3.25) -- (4.5,1);
\node[] at (4.5,1) {$\times$};

\node[rotate=\iangle]  at (3.25,10.25) {\textbf{Object}};;
\node[rotate=\iangle] at (4.75,6.75) {\textbf{Robot}};
\node[rotate=\iangle]  at (1.75,8) {$\phi(\mathbf{q})$};
\node[color = green] at (6.5,10.5) { \textbf{Goal}};
\node[color = green] at (5,1) { \textbf{Goal}};
\end{tikzpicture}
}
    \caption{The gap function is defined as the signed distance between the origin of frame $\{E\}$ and a virtual plane that separates the object on one side from the robot on the other side. In that way, $\phi(\mathbf{q})$ becomes negative whenever the end-effector penetrates the object. This plane contains the origin of frame $\{O\}$ and has $_\mathcal{I}\mathbf{x}_o$ as its normal vector, so it is also the $y_oz_o$-plane}
    \label{fig:DistanceFunction}
\end{figure} 
A generalized gap function $\phi(\mathbf{q})$ is introduced for the ANYpulator-Object problem, while further adding geometry-dependent constraints to the NLP. The minimum signed distance between the robot's end-effector and any convex-shaped object is defined as the projection of the position vector $_O \mathbf{r}_{eo}$ onto the object-frame's $x$-axis (refer to~\cref{fig:DistanceFunction}). Therefore,
\begin{equation}
\begin{split}
    \phi(\mathbf{q}) &= _O r_{eo_x} \\ 
    &= \begin{bmatrix}1 & 0 & 0\end{bmatrix}\cdot \left(\mathbf{R}^{\mathcal{I}}_O\right)^T \cdot \left( _{\mathcal{I}}\mathbf{r}_{o} - \ _{\mathcal{I}}\mathbf{r}_{e}\right)
\end{split}
\end{equation} \\
where $\{O\}$ is a body-attached frame (rotates and translates with the object), and its $x$-axis is directed along the assigned direction of motion, while $\mathbf{R}^{\mathcal{I}}_O$ represents the orientation of frame $\{O\}$ with respect to frame $\{\mathcal{I}\}$.   
In order to ensure that the robot's end-effector actually comes in contact with the block-face -- and not only with the separating virtual plane -- the following nonlinear inequality conditions are added
\begin{equation} \label{eq:6.4}
        \begin{cases}
        \lambda\cdot\left(h_b- \ _{\mathcal{I}}r_{e_z}\right) \geq 0 \\
        \lambda\cdot\left(\left(\dfrac{w_b}{2}\right)^2- \ _{O}r_{eo_y}^2\right) \geq 0 
        \end{cases}
\end{equation}
where $w_b$ and $h_b$ are the block-face's width and height, respectively. It is assumed here that the block's bottom surface and the robot base are at the same level. \\ 
    
\textit{Other considerations:} An extra path constraint is introduced in order to ensure that the resulting optimal motion plan in joint space does not lead to collisions between the end-effector and the ground plane. Furthermore, the gradients of both the objective function and the nonlinear inequality constraints are provided analytically to reduce the computational load of the program. 
    
\subsection{Control Policy}
The output obtained from solving the optimization problem formulated in~\cref{subsec:NLP} can be visualized to verify the satisfaction of the complementarity conditions, and to validate the attainment of the desired manipulation task. Before testing the motion plan on the real setup, we perform simulations within the \textit{Gazebo} simulation environment \cite{gazebo}, with Open Dynamics Engine (ODE) as its underlying physics engine (it consists of a rigid body dynamics module in addition to a collision detection engine). 

The shift from pure visualization to either simulation or experimentation requires the introduction of a closed-loop control policy that is capable of tracking the optimal joint states. A simple yet robust approach is adopted in this work, where the optimal control problem is solved only once; then the resulting open-loop torque sequence is added as a feedforward term to a feedback term acting on the deviations from the optimal position and velocity trajectories. Hence, the control law is derived as follows, 
\begin{equation} \label{controlLaw}
    \tau_i = \underbrace{\tau_{opt_i}}_{\text{feedforward term}} + \ \ \ \underbrace{k_{p_i}\left(q_{opt_i}-q_i\right) + k_{d_i}\left(\dot{q}_{opt_i}-\dot{q_i}\right)}_{\text{feedback term}}
\end{equation} \\
where the index $i$ refers to an element from the set of actuated degrees of freedom; while $k_{p_i}$ and $k_{d_i}$ are the scalar feedback PD-gains. $\tau_{opt}(t)$ is obtained by applying a zero-order-hold on the sequence $\tau_{opt_k}$, while $q_{opt}(t)$ and $\dot{q}_{opt}(t)$ by linearly interpolating the grid-point-values $q_{opt_k}$ and $\dot{q}_{opt_k}$ respectively.
 
A C++ class structure is developed -- along with the use of libraries and tools provided by the Robot Operating System (ROS) middleware -- to send commands to the ANYdrives with an update frequency of ${1000 \ \text{Hz}}$. These commands include the optimal feedforward torques, reference joint positions, and reference joint velocities. The PD gains are manually tuned with the following values during simulation: $\mathbf{K}_p = [20 \ \ 50 \ \ 50\ \ 50 \ \ 50]^T$, $\mathbf{K}_d = [10 \ \ 15 \ \ 15 \ \ 1 \ \ 1]^T$; and during experimentation: $\mathbf{K}_p = [110 \ \ 120 \ \ 100 \ \ 80 \ \ 80]^T$, $\mathbf{K}_d = [0.2 \ \ 0.1 \ \ 0.05 \ \ 0.15 \ \ 0.05]^T$.

\section{Experimental Setup}
To have a somewhat accurate placement of the initial block pose, its starting position and orientation are provided as target references for the robot's tool frame (frame $\{E\}$), which are then tracked with an inverse kinematics controller, to place the block accordingly.
Moreover, for the sake of comparing the contact forces generated while performing our experiments to the forces predicted by the optimization, a force/torque sensor is mounted on the robot end-effector. 

\begin{figure}[b!]
\hspace*{-0.2cm}
\makebox[0.5\textwidth][c]{   
\centering
\includegraphics[scale=0.19]{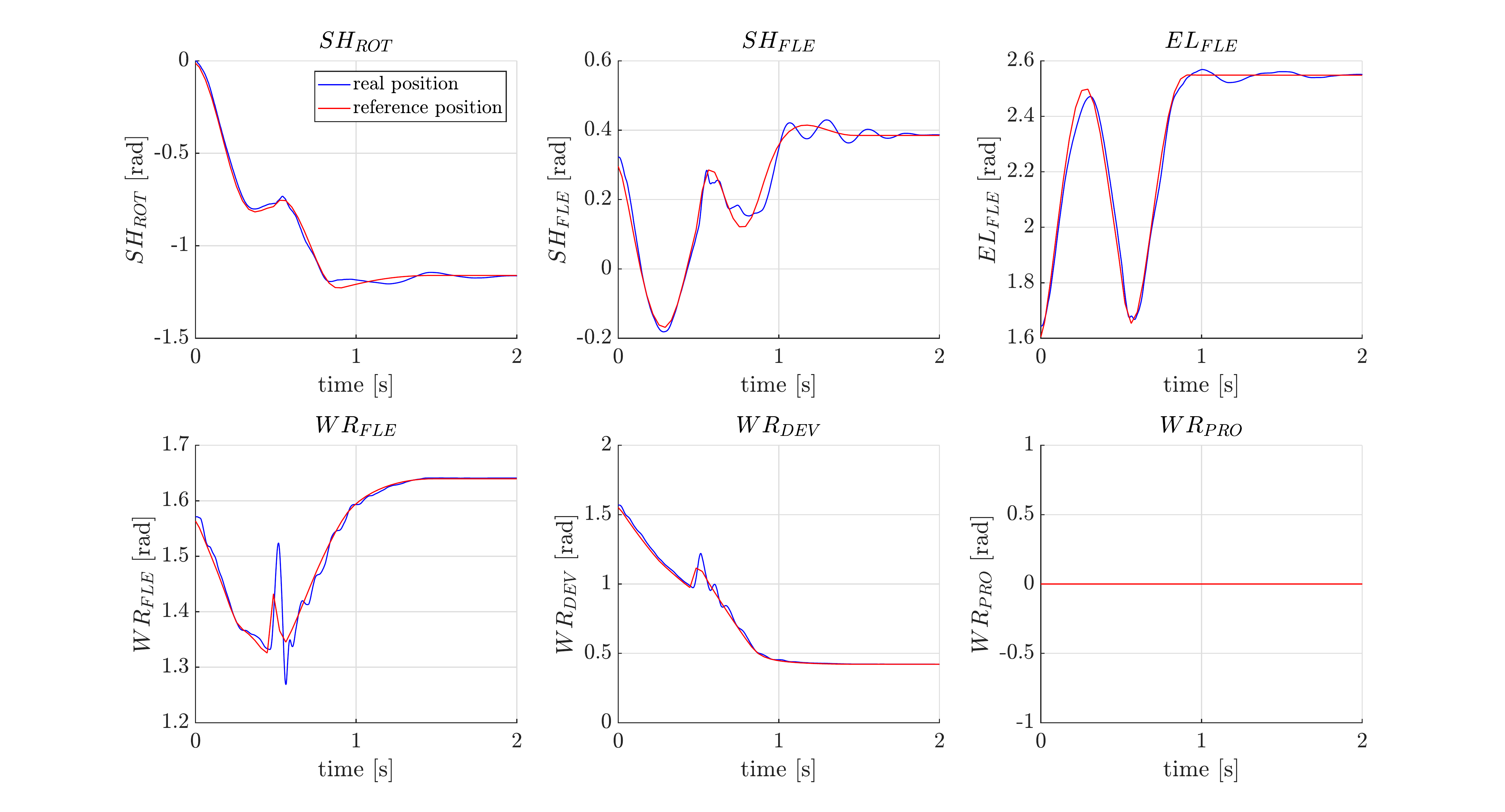}              
}  
\caption{Joint position tracking of optimal references by the real ANYpulator setup, for the first experiment}      
\label{fig:jointPositions}
\end{figure}
\begin{figure}[b!]
\hspace*{-0.2cm}
\makebox[0.5\textwidth][c]{
\centering
\includegraphics[scale=0.19]{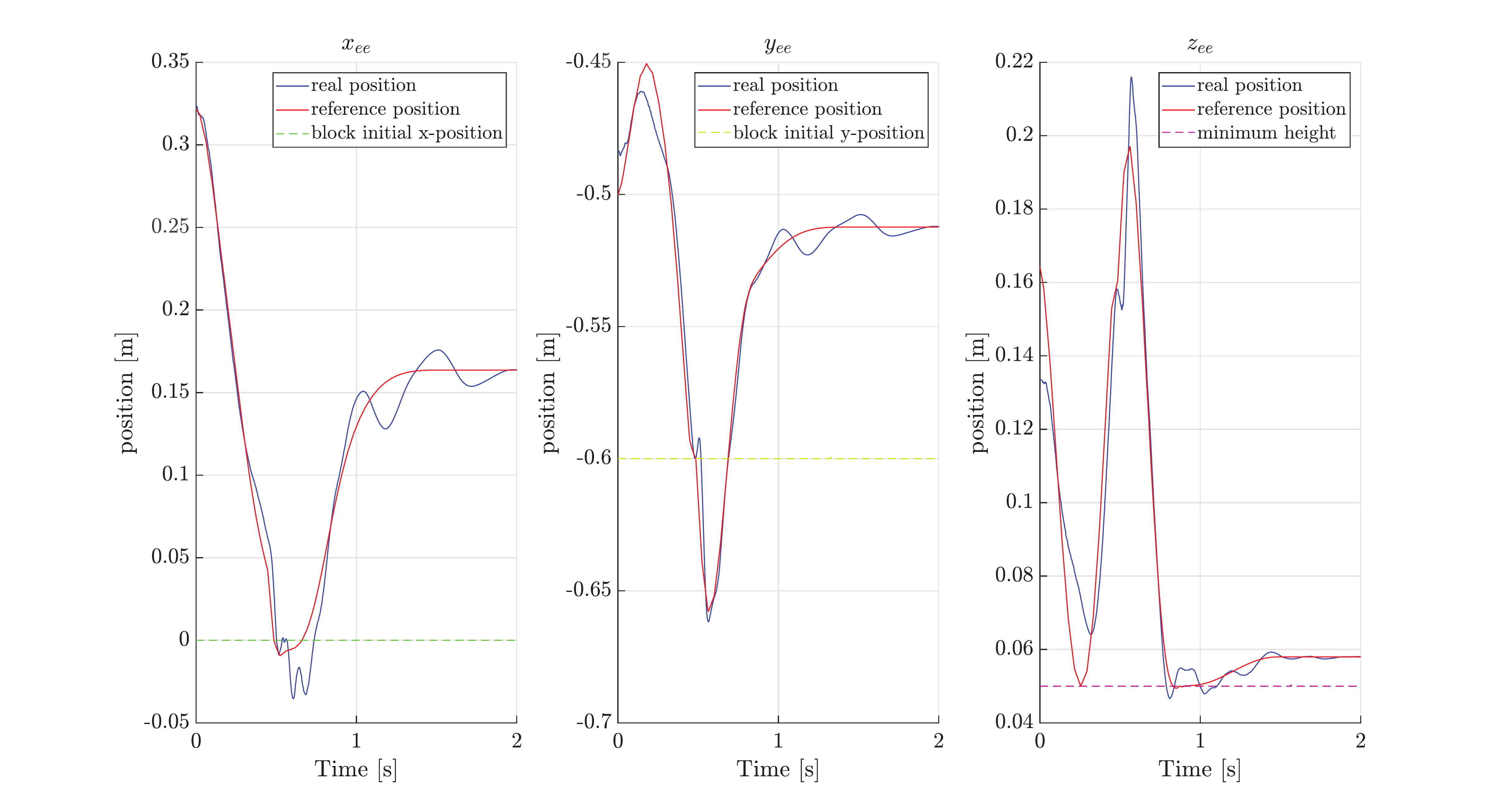}              
}  
\caption{ANYpulator end-effector tool position (in cartesian coordinates) tracking optimal reference trajectories, given by forward kinematics (Experiment 1)}      
\label{fig:eePositions}
\end{figure}

Several test runs are performed on variations of the dynamic block-pushing problem (i.e., different robot initial conditions, different block starting configurations, and different final desired positions for the block), both in simulation and hardware experimentation. We mainly emphasize three experiments that are tested five times each, and then analyze the results and draw conclusions from them. These experiments involve the same initial block pose and robot configuration; however, the desired displacement of the block varies among them, with experiments 1, 2, and 3 corresponding to displacements of ${0.7 \ \text{m}}$, ${0.6 \ \text{m}}$, and ${0.4 \ \text{m}}$ respectively. 
\begin{figure*}[t!]
\centering
\vspace*{-0.2cm}
\begin{tikzpicture}
\setlength{\fboxsep}{0pt}%
\setlength{\fboxrule}{1pt}%
 \node[] (image1) at (0,0) {\includegraphics[width=4.3cm,keepaspectratio]{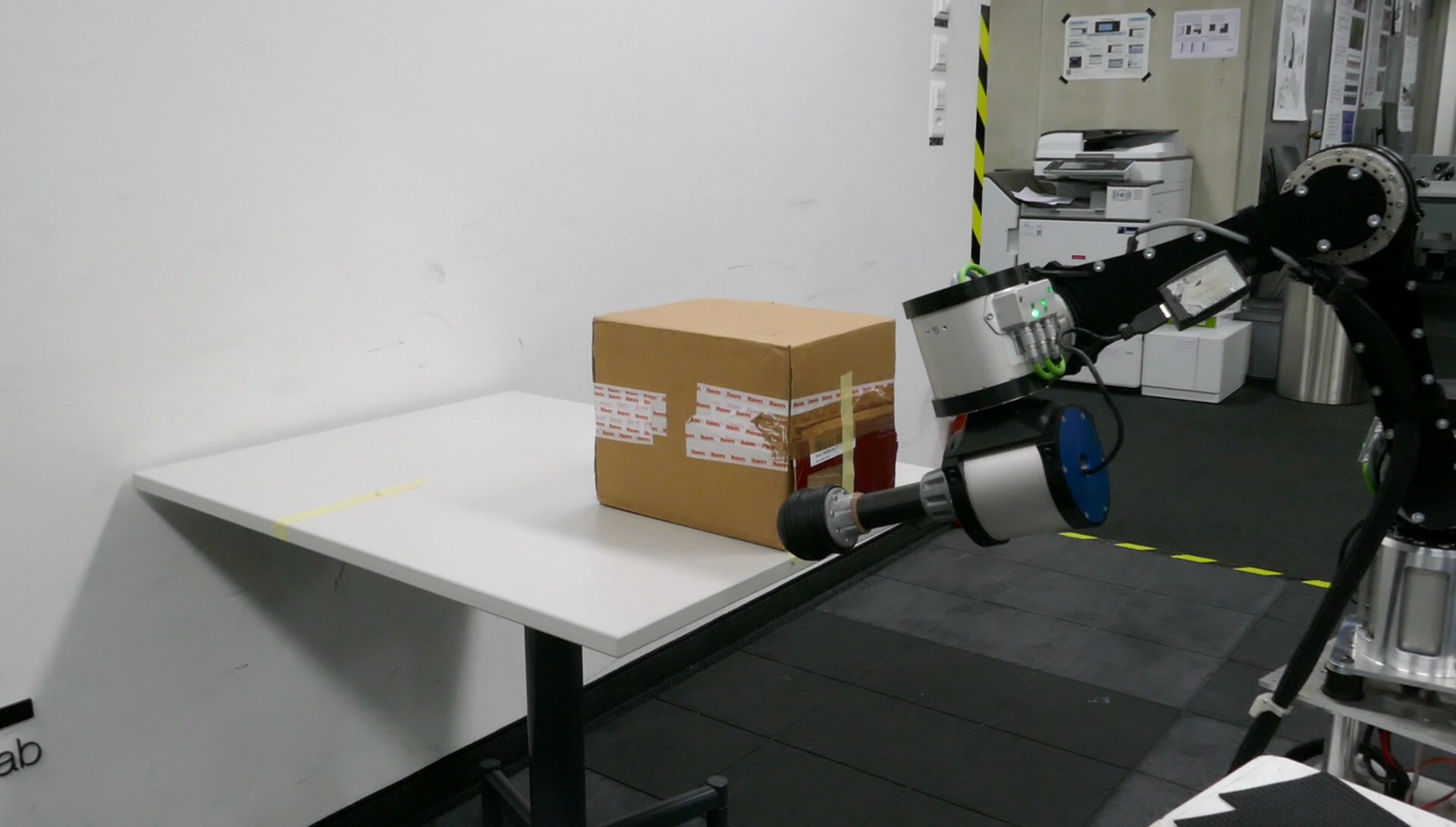}};
 \node[rectangle, minimum height =0.75cm, minimum width = 0.5cm,yshift = -0.5cm,xshift = 0.5cm] at (image1.north west) {1};
 \node[] (image2) at (4.4,0) {\includegraphics[width=4.3cm,keepaspectratio]{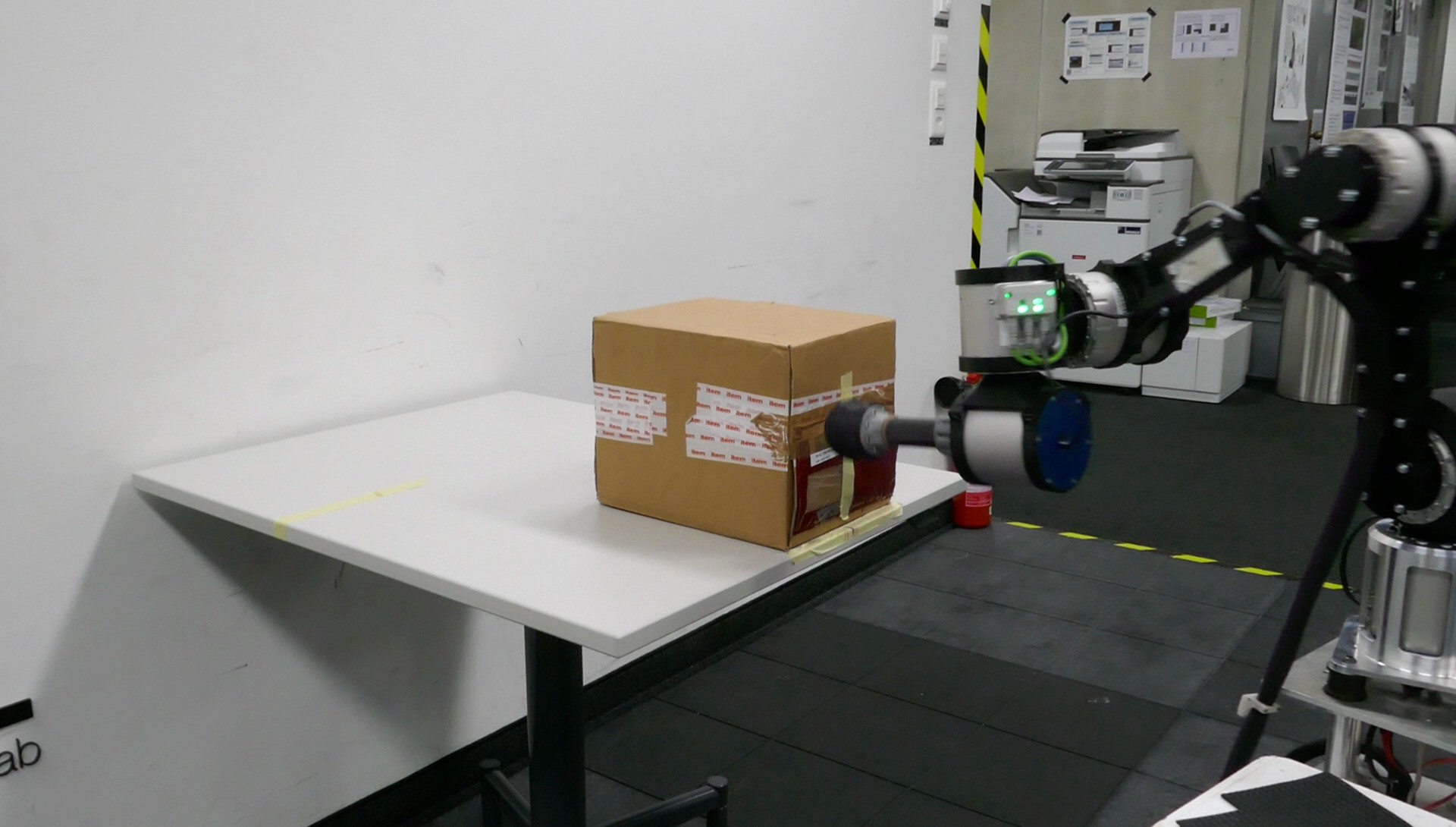}};
  \node[rectangle, minimum height =0.75cm, minimum width = 0.5cm,yshift = -0.5cm,xshift = 0.5cm] at (image2.north west) {2};
 \node[] (image3) at (8.8,0) {\includegraphics[width=4.3cm,keepaspectratio]{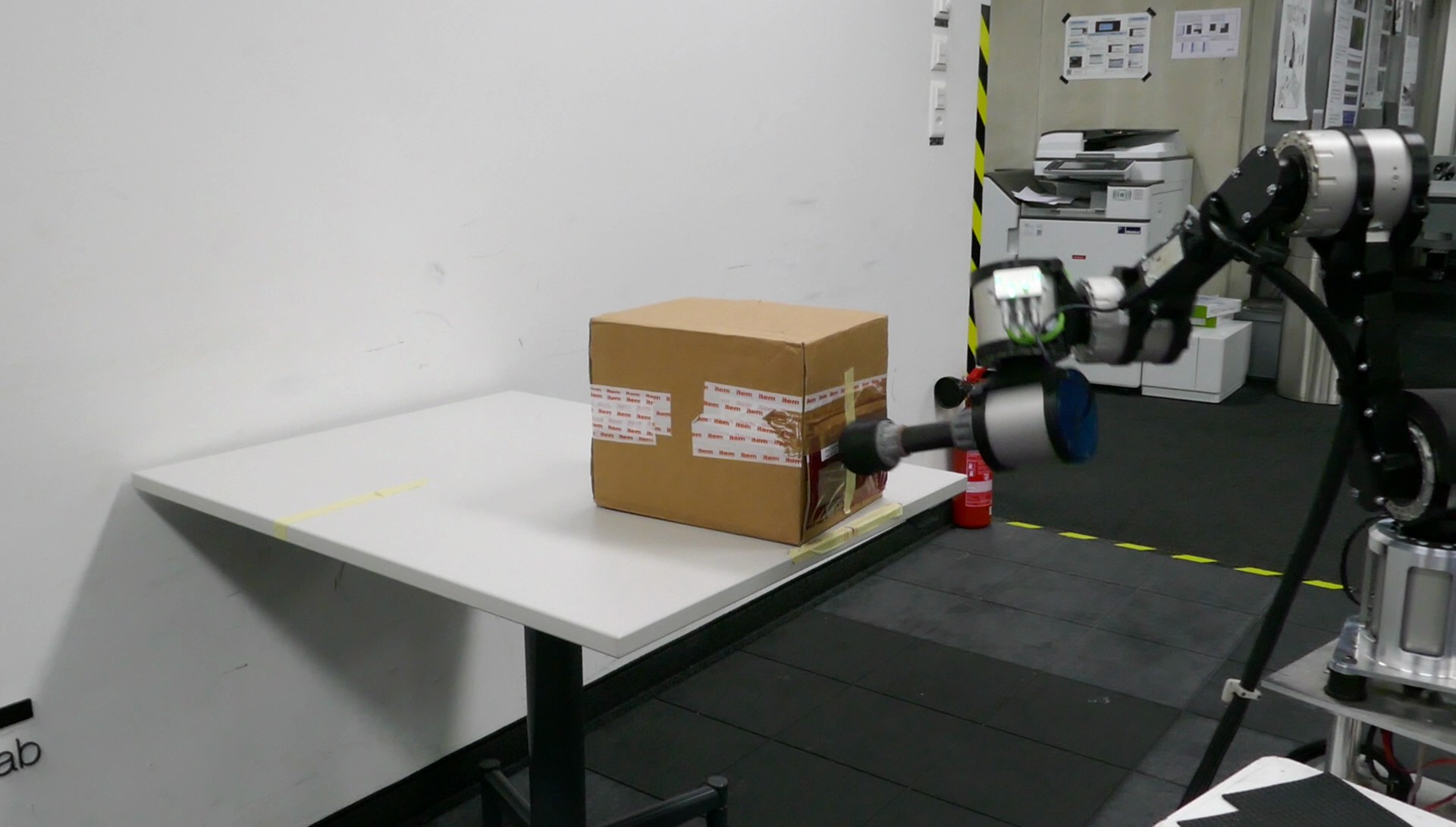}};
  \node[rectangle, minimum height =0.75cm, minimum width = 0.5cm,yshift = -0.5cm,xshift = 0.5cm] at (image3.north west) {3};
 \node[] (image4) at (13.2,0) {\includegraphics[width=4.3cm,keepaspectratio]{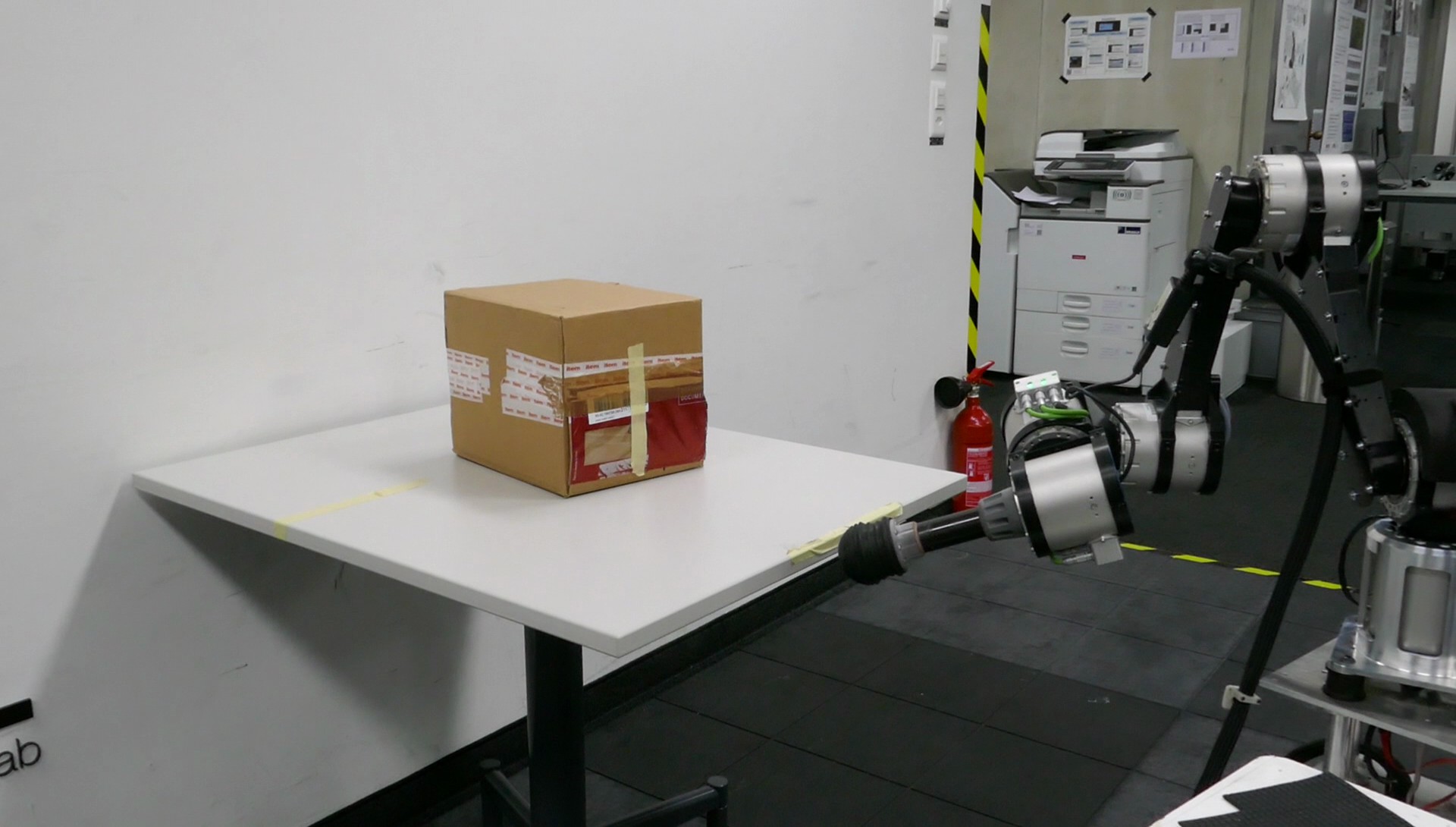}};
  \node[rectangle, minimum height =0.75cm, minimum width = 0.5cm,yshift = -0.5cm,xshift = 0.5cm] at (image4.north west) {4};
\end{tikzpicture}
\caption{Snapshots from Experiment 3: Desired block displacement of $0.4 \ \text{m}$. See video for additional results (\url{https://youtu.be/YHje00XY4go})}
\label{fig:experiments}
\end{figure*}
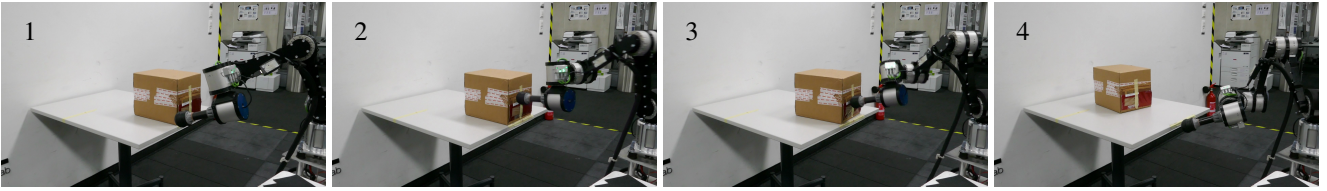
\begin{figure}[t!]
  \makebox[0.5\textwidth][c]{
\minipage{0.25\textwidth}
\includegraphics[height = 3.6 cm, keepaspectratio]{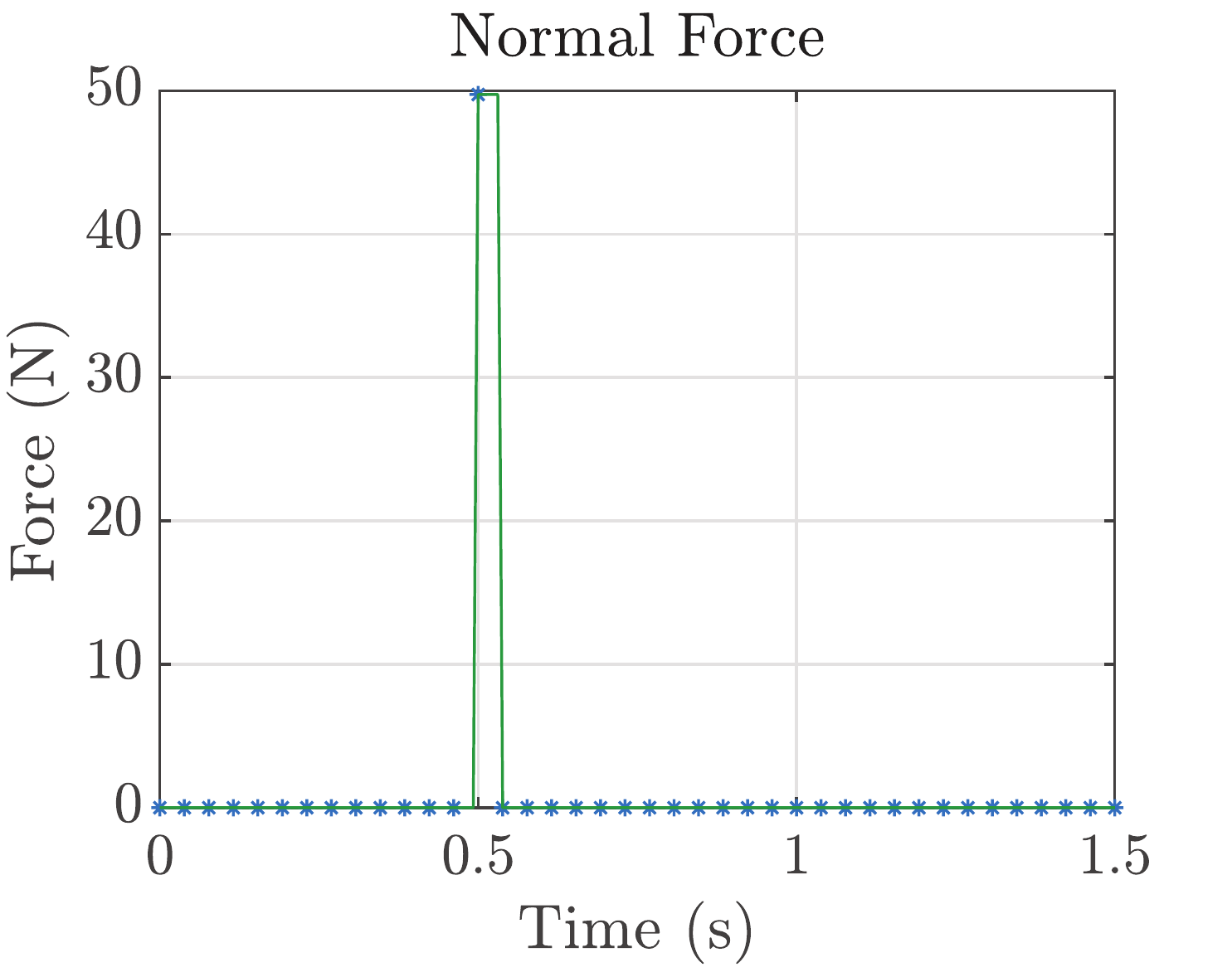}
\endminipage
\minipage{0.25\textwidth}
\includegraphics[height = 3.6 cm, keepaspectratio]{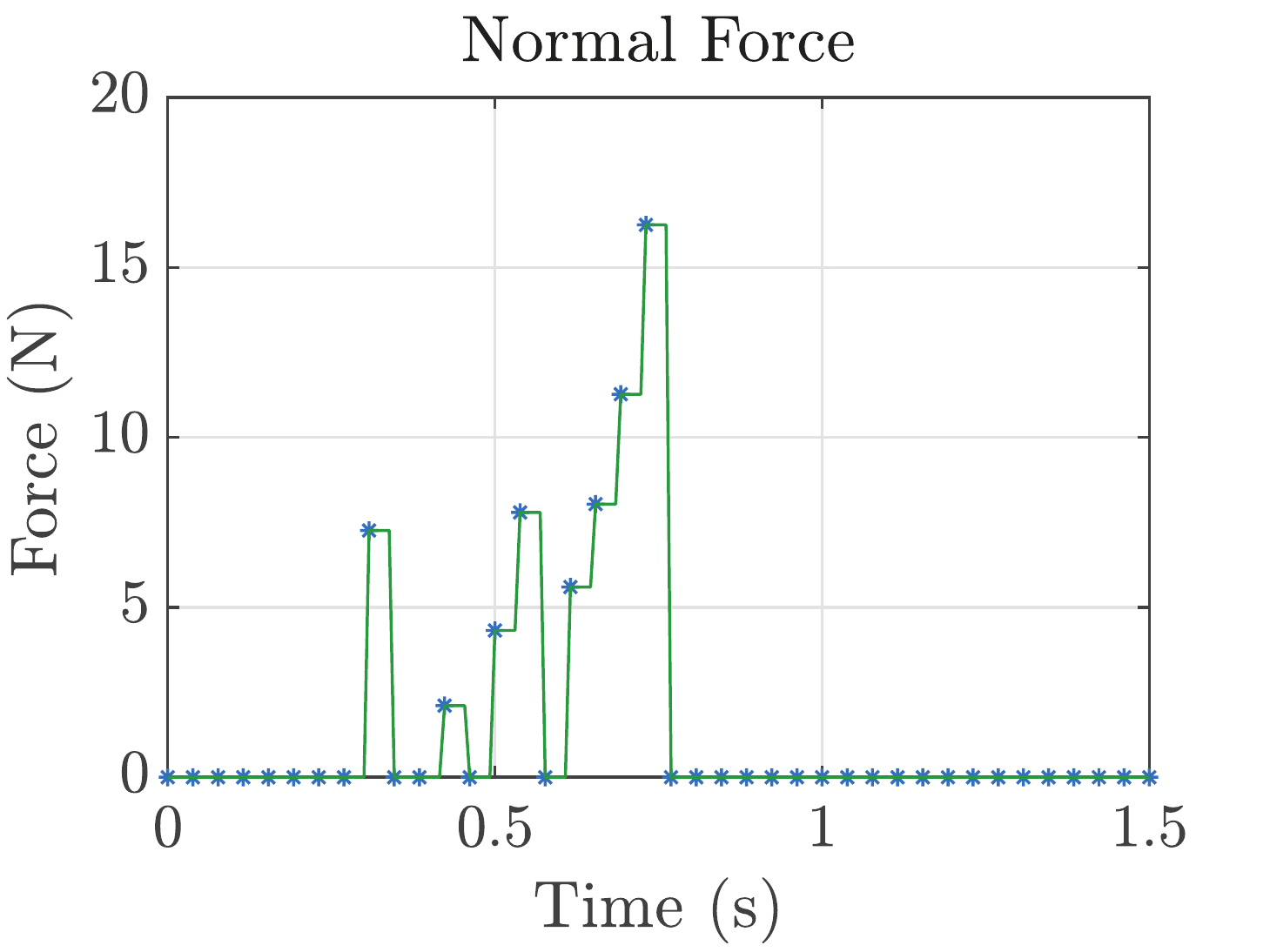}
\endminipage}
\caption{Normal contact force sequence discovered by the optimization, for the first experiment \textit{(left)} and second experiment \textit{(right)}}
\label{ForcePlots2}
\end{figure}
\begin{figure}[t!]
  \makebox[0.5\textwidth][c]{
\minipage{0.25\textwidth}
\includegraphics[height = 3.6 cm, keepaspectratio]{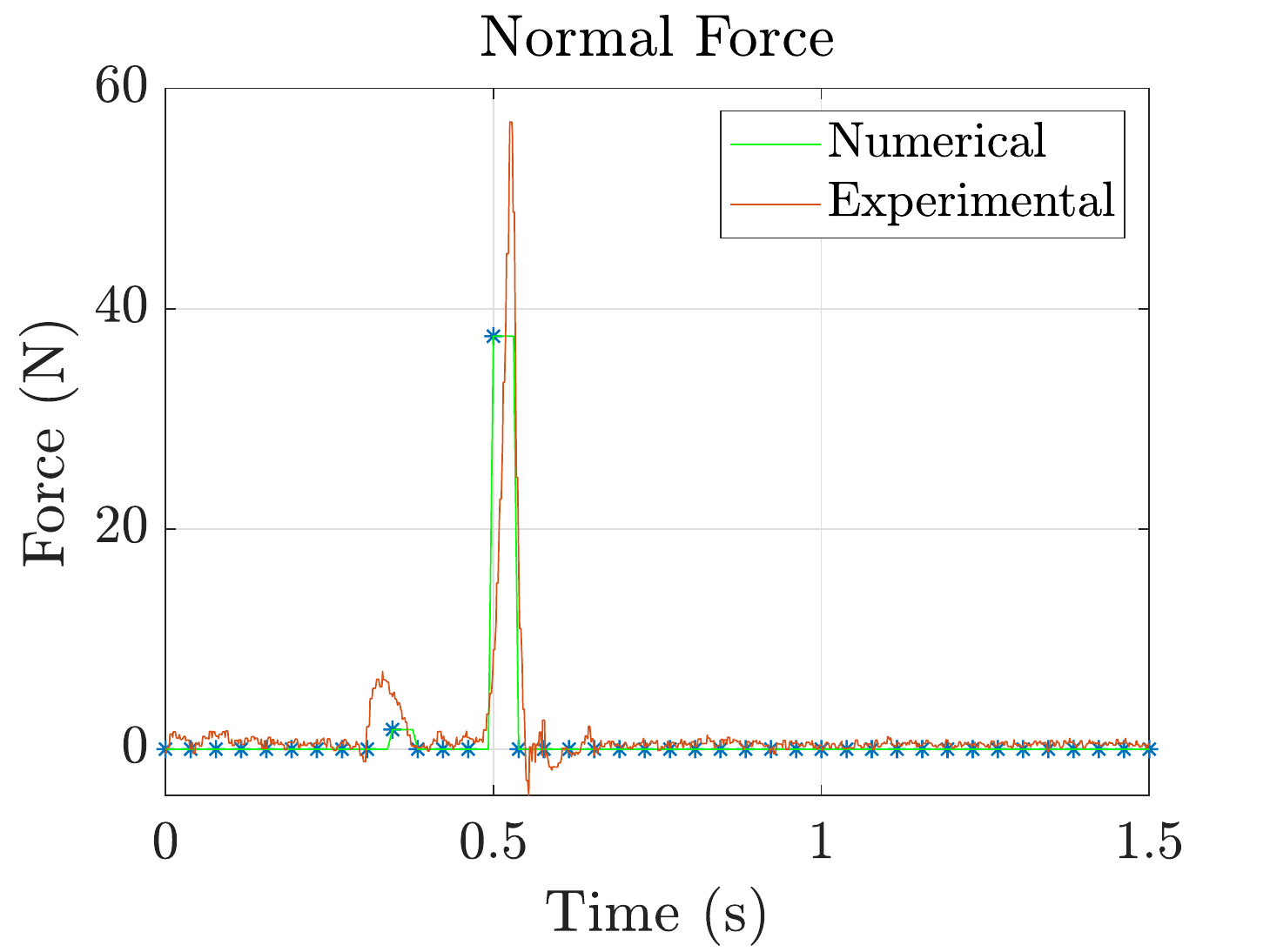}
\endminipage
\minipage{0.25\textwidth}
\includegraphics[height = 3.6 cm, keepaspectratio]{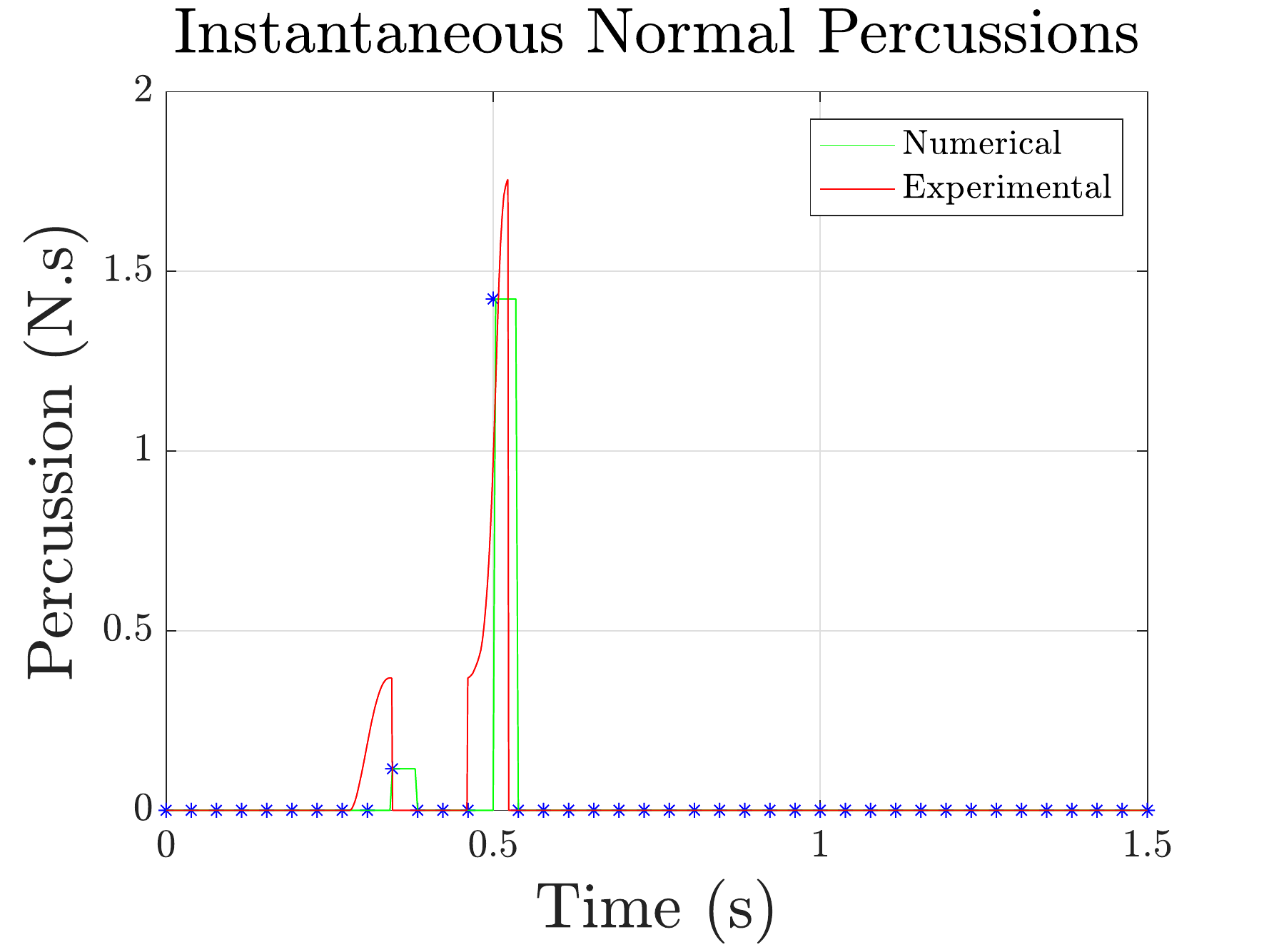}
\endminipage}
\caption{Optimal vs. Measured normal contact forces \textit{(left)} and equivalent normal percussions \textit{(right)} for the third experiment}
\label{ForcePlots}
\end{figure}
\section{Results}
The formulated CIO program needed to solve the ANYpulator-Block task consists of 896 optimization variables, 819 equality constraints, and 1955 inequality constraints. On average, it takes the solver around ${800 \ \text{ms}}$ to converge to a local minimum, but this value can vary depending on the utilized initial guess, and on the manipulation-problem specifications.
Proper tracking of the optimal reference joint-positions as well as the corresponding end-effector references, is demonstrated for the first experiment in~\cref{fig:jointPositions} and~\cref{fig:eePositions}, respectively. It could be noted from the cartesian coordinates of the tool that the robot indeed reaches the block's initial position and tends to sustain the closed contact (the push occurs along the negative $y$-axis of the inertial frame) as much as specified by the optimal plan, while also maintaining a tool height greater than ${5 \ \text{cm}}$ throughout the full time-horizon. Moreover, the effect of the impact-event can be directly perceived in these plots as well, especially in the $WR_{FLE}$ joint, where it is evident that a disturbance is introduced after around 0.5 seconds. Indeed, ~\cref{ForcePlots2} shows this to adhere with the optimal force plan, which involves a contact event occuring as a single impulsive push with a force of about ${\lambda = 50 \ \text{N}}$ acting at ${t=0.5 \ \text{s}}$ for ${\Delta t = 0.0387 \ \text{s}}$. More interestingly, in the second experiment, we perceive a contact event that lasts for approximately ${8\cdot \Delta t = 0.3096 \ \text{s}}$ (see~\cref{ForcePlots2}), while in the last one we encounter two separate contacts -- one with a small magnitude and the other with a large one, held over a single time step each -- as shown in~\cref{ForcePlots}. The generated normal force $\lambda$ is shown to decrease with the assigned displacement along the three experiments. 
\begin{figure}[t!]
    \centering
{%
\setlength{\fboxsep}{0pt}%
\setlength{\fboxrule}{0.1pt}%
\includegraphics[scale=0.5]{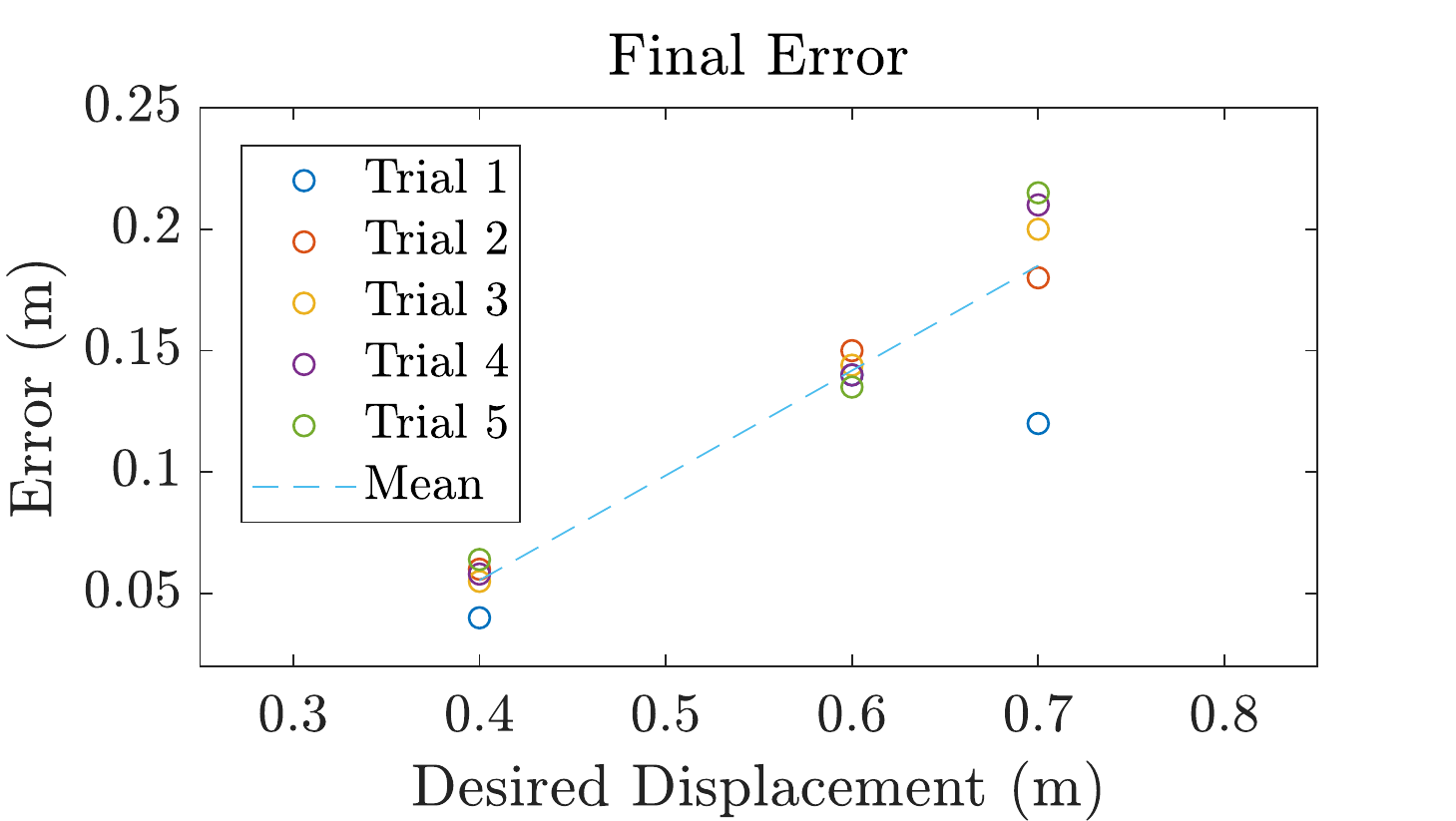}%
}%
    \caption{Observed error (in absolute value) between the block's final position and the desired one throughout the three experiments, which are performed five times each}
    \label{errors}
\end{figure}
The resulting optimal trajectories associated with the contact force, indicate the richness of the CIO approach in general, as it yields contact schedules that could not be easily considered or planned for, within a multi-phase framework. In addition to that, the dynamic feasibility of the applied method is demonstrated in three different settings, by observing the properly-tracked motion plan, and by noting that the predicted force sequence is indeed compatible with the force sensor readings (see~\cref{ForcePlots}), which is significant given that our control law does not include any closed-loop force tracking term. These two factors are fundamental for achieving the assigned manipulation task at hand, as presented for the third experiment in~\cref{fig:experiments}.

One should keep in mind that eventhough the optimal and measured force peak magnitudes in the contact force plot are not equivalent, the resulting normal percussions turn out to be reasonably similar. It is also important to recall that each experiment is repeated 5 times in order to demonstrate the method's repeatability and reliability. Indeed this is shown to be true, since the same results come up repeatedly, for various runs of the same experiment. To add an accuracy-based evaluation of the method, we refer to the plot shown in~\cref{errors}, which indicates the final error (in absolute value) between the real block position and the desired one. First of all, it could be noticed that in all of our experiments, the block always travels a distance that is less than the expected one. This is likely related to the inaccuracy in our dynamic model in general and specifically the assumed static friction model, which seems to have underestimated the true frictional effects. Secondly, we observe a trend indicating that the higher the desired block displacement, the more erroneous is the outcome. This could be partly related to the fact that the higher the demanded displacement, the larger the generated contact force; and since there could be a slight deviation from the supposed contact location (the central line of the box), then this would lead to the block's rotation -- which is proportional to the contact force's magnitude -- along with its translation. As a result, it loses a part of its total kinetic energy, thus having less energy remaining for the translational motion. Moreover, the neglected frictional forces acting in the robot-to-object tangent contact plane may as well lead to the block's undesired rotation. Even if the peak in the optimal force plan does not turn out to be larger (i.e., for a different local minimum) when assigning a greater displacement, contact is now established earlier, and for a longer time. Otherwise, a bigger displacement could not be attained by the end of the same time horizon. Consequently, the inaccuracy increases due to the greater accumulated integration error associated with the block dynamics.  
\section{CONCLUSIONS}
The aim of this work has been to make use of a contact-implicit optimization approach to tackle a variety of dynamic pushing problems, performed by the ANYpulator robot-arm, with particular emphasis on the block-task. This was done by formulating the manipulation task as a multi-stage program, thus making it possible to integrate it within the \textit{FORCES Pro} framework. The NLP is solved efficiently, and yields both optimal state and input trajectories, which, when used as components of a simple closed-loop control scheme, result in reasonable tracking of the reference motion and contact force. The resulting optimal trajectories indicate the richness of this approach in terms of its ability to discover different contact-schedules, while the experimental findings suggest the effectiveness of the control technique in terms of its simplicity, the achieved dynamic feasibility, and repeatability. As for the task-satisfaction accuracy of the method, it has been shown to be adequate, but decreases with the assigned block displacement. 

There are several directions for potential future work and considerations, most notably is the further improvement of the method's computational efficiency, in order to incorporate it within an NMPC framework. Possible ways of achieving that could be by either introducing effective warm-starting techniques for the currently adopted interior-point algorithm, or by exploring different algorithms in the first place, that could potentially enhance efficiency when solving the CIO program in an online-fashion. It would then be compelling to test the NMPC version as an integrated motion planner and tracking controller within other contact-rich domains, such as locomotion, where multiple contacting bodies are involved, and contacts are generally sustained for longer periods of time.

\addtolength{\textheight}{0cm}   

\bibliographystyle{IEEEtran}
\bibliography{IEEEabrv,mybibfile}

\end{document}